\documentclass{article}

    \PassOptionsToPackage{numbers, compress}{natbib}
    \usepackage[preprint]{neurips_2023}

\usepackage[utf8]{inputenc} % allow utf-8 input
\usepackage[T1]{fontenc}    % use 8-bit T1 fonts
\usepackage{hyperref}       % hyperlinks
\usepackage{url}            % simple URL typesetting
\usepackage{booktabs}       % professional-quality tables
\usepackage{amsfonts}       % blackboard math symbols
\usepackage{amsmath}
\usepackage{nicefrac}       % compact symbols for 1/2, etc.
\usepackage{microtype}      % microtypography
\usepackage[usenames,dvipsnames]{xcolor} %colors
\usepackage{multirow}
\usepackage{tabularx}
\usepackage{booktabs}% for better rules in the table
\usepackage{comment}

\usepackage{cleveref}
\newcommand\myshade{85}
\colorlet{mylinkcolor}{YellowGreen}
\colorlet{mycitecolor}{OliveGreen}
\colorlet{myurlcolor}{SpringGreen}

\hypersetup{
  linkcolor  = mylinkcolor!\myshade!black,
  citecolor  = mycitecolor!\myshade!black,
  urlcolor   = myurlcolor!\myshade!black,
  colorlinks = true,
}

\usepackage[pdftex]{graphicx}

\usepackage{pifont}% http://ctan.org/pkg/pifont
\newcommand{\cmark}{\ding{51}}%
\newcommand{\xmark}{\ding{55}}%

\DeclareMathOperator*{\argmin}{arg\,min} % thin space, limits underneath in displays

\title{TSPP: A Unified Benchmarking Tool for Time-series Forecasting}

\makeatletter
\newcommand{\mpm}{\mathbin{\mathpalette\@mpm\relax}}
\newcommand{\@mpm}[2]{\ooalign{%
  \raisebox{.1\height}{$#1+$}\cr
  \smash{\raisebox{-.6\height}{$#1-$}}\cr}}
\makeatother

\author{%
  Jan Bączek \\
  \And
  Dmytro Zhylko \\
   \And
  Gilberto Titericz \\
   \And
  Sajad Darabi \\
   \And
  Jean-Francois Puget \\
   \And
  Izzy Putterman \\
   \And
  Dawid Majchrowski \\
   \And
  Anmol Gupta \\
   \And
  Kyle Kranen \\
   \And
  Pawel Morkisz \\
}

\begin{document}

\maketitle
\vspace{-3em}
\begin{center}
     \textbf{NVIDIA}
\end{center} 
\vspace{1em}

\begin{abstract}
    While machine learning has witnessed significant advancements, the emphasis has largely been on data acquisition and model creation. However, achieving a comprehensive assessment of machine learning solutions in real-world settings necessitates standardization throughout the entire pipeline. This need is particularly acute in time series forecasting, where diverse settings impede meaningful comparisons between various methods. To bridge this gap, we propose a unified benchmarking framework that exposes the crucial modelling and machine learning decisions involved in developing time series forecasting models. This framework fosters seamless integration of models and datasets, aiding both practitioners and researchers in their development efforts. We benchmark recently proposed models within this framework, demonstrating that carefully implemented deep learning models with minimal effort can rival gradient-boosting decision trees requiring extensive feature engineering and expert knowledge. The framework is accessible on \href{https://github.com/NVIDIA/DeepLearningExamples/tree/master/Tools/PyTorch/TimeSeriesPredictionPlatform}{here}.
\end{abstract}

\section{Introduction}
Time series forecasting plays a crucial role in diverse fields like finance, weather prediction, and demand forecasting. Accurate forecasts empower businesses to optimize decision-making, enhance operations, and improve overall efficiency. However, the inherent complexities within time series data, including trends, noise, missing values, and evolving relationships between variables, present significant challenges to achieving accurate forecasts.

To address the challenges of time series forecasting, a diverse landscape of methods has emerged. Gradient boosting machines (GBM) \citep{GBM} have become popular in Kaggle competitions due to their effectiveness, but require substantial feature engineering effort. While promising, deep learning models are less frequently used independently, primarily due to data limitations. The NN3 competition aimed to showcase the viability of neural networks (NNs) for time series forecasting, but initial results published in 2011 \citep{NN3} were disappointing, highlighting the dominance of statistical methods. This was further confirmed by the 2017 Web Traffic Time Series Forecasting competition \citep{wikitraffic}, where machine learning methods performed better with abundant and homogeneous data. Similarly, the renowned M4 competition \citep{m4} demonstrated the superiority of statistical approaches on heterogeneous data. Despite this trend, an ensemble combining an RNN and a statistical model won the competition \citep{m4_winner}, although it was unclear whether this signaled a shift towards deep learning. The successor M5 competition \citep{m5}, utilizing real-world sales data from Walmart, witnessed the dominance of gradient boosting machines, with the second-place solution leveraging an ensemble of GBM and a neural network \citep{m5_2nd_place,nbeats}. Suggesting a potential shift towards deep learning architectures becoming an integral part of the time series forecasting pipeline.

The rapid evolution of diverse time series forecasting approaches has made it difficult to compare and establish consensus on what constitutes a high-performing deep learning model. While efforts exist to standardize datasets for benchmarking these models \citep{monash}, comparisons often overlook crucial aspects across the entire machine learning lifecycle, leading to ambiguous results that offer limited guidance for model improvement. For instance, established techniques for training neural networks, such as curriculum learning, exponential moving average, and drift estimation, are readily available but largely ignored in the current literature. These studies frequently adopt a \texttt{model-centric} perspective, focusing solely on the forecasting algorithm itself. Conversely, our work aims to address this gap by comprehensively considering all moving parts within the machine learning lifecycle. To achieve this, we propose a unified benchmarking framework that standardizes the evaluation of forecasting models. This framework facilitates the implementation and comparison of the most influential time series forecasting models, enabling a more comprehensive and objective assessment of their performance.

Our contribution are as follows: \textbf{1)} We develop and open-source a standardized framework for time series forecasting benchmarking. The framework has modular components that enable fast and easy integration of datasets, models, and training techniques. \textbf{2}) Using the proposed framework, recent popular deep learning models for time-series forecasting are re-implemented and improved upon, assesing their preformance on commonly used datasets across these methods. \textbf{3)} Various observations are outlined from the extensive computational resources allocated to tuning these models and comparing them. The complete code and detailed reports for our experiments \href{https://github.com/NVIDIA/DeepLearningExamples/tree/master/Tools/PyTorch/TimeSeriesPredictionPlatform}{here}.

\section{Related work}

The majority of research in time series forecasting has revolved around competitions \citep{brief_hist}, where participants utilize diverse methodologies. These methods fall into three main categories: 1) statistical methods, such as ARIMA \citep{arima}, 2) classical machine learning algorithms, including support vector machines \citep{hearst1998support}, gradient boosting machines \citep{chen2015xgboost}, and 3) neural network-based approaches, exemplified by TFT \citep{tft} (a transformer-based model), NHITS \& NBEATS \citep{nhits, nbeats} (which utilize a hierarchical approach with multi-layer perceptrons), and MTGNN \citep{mtgnn} (another model leveraging the power of Graph Neural Networks (GNNs) by incorporating graph structure). However, the use of datasets with diverse settings in the methods published hinders accurate performance evaluation and makes it difficult to identify a clear frontrunner.

Despite efforts to standardize benchmarking practices, the variability of datasets across methods hinders meaningful comparisons. In \citep{forecast5010017}, authors highlight the limitations of dataset comparisons due to diverse features like size, number of features, and noise level. Similar observations are echoed by benchmarks like Monash Time Series Forecasting Benchmark\citep{monash}, which curates datasets specifically for model comparison. However, initiatives like \citep{januschowski2020criteria} focus on classifying model operating modes (univariate/multivariate, local/global) as the basis for comparison. In contrast, \citep{lara2021experimental} takes a model-centric approach, exploring architectural changes in neural networks. Their findings suggest the absence of a universal best architecture, highlighting the data-dependence of model performance. While these studies acknowledge the lack of standardization, their focus is either \textit{model-centric} or  \textit{data-centric} which hinders a comprehensive understanding of the interplay between the two. Establishing a standardized framework that considers both aspects simultaneously is crucial for fostering meaningful comparisons and advancements in time series forecasting.

While deep learning models have gained popularity, recent works challenge their effectiveness against simpler, more interpretable methods \citep{dowereally,zeng2022transformers,liu2023we}. However, we argue that deep learning models haven't received their due credit. With careful consideration throughout the entire machine learning life-cycle and appropriate optimization techniques, we believe these models can achieve further improvements beyond their initially claimed capabilities.

\section{Time-series Forecasting Framework}

\begin{figure}[h]
  \centering
  \includegraphics[width=0.8\linewidth]{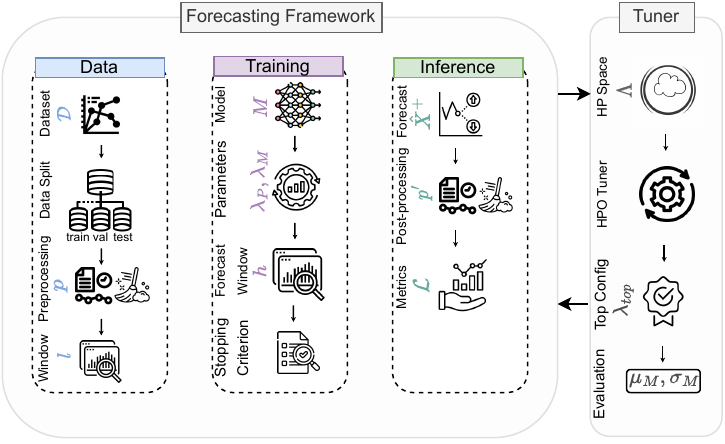} 
  \caption{An overview of the pipeline standardization considered to guide model development and comparison. The framework is comprised of three main components, data, training, inference, which a tuner has access to for adequate exploration to ensure reliable model training.}
  \label{fig:framework_overview}
\end{figure}

Time series encompass a sequence of data points indexed by time, and forecasting is the process of predicting future values from past historical data. Albeit the task can be defined simply in words,  the machine learning life cycle for such models are quite complex and an \textit{end-to-end machine learning pipeline} covering this from data curation to deployment is highly needed.  An overview of the framework proposed is depicted in Figure \ref{fig:framework_overview} which comprises of the following components:
\begin{itemize}
    \item \textbf{Data}: Decisions made for improving the data and shaping it for model training. This includes data selection, curation and cleaning.
    \item \textbf{Training}: Decisions made with respect to the input data and model in order to improve downstream performance.  This includes model design tailored to input data, optimization, and selection criterion.
    \item \textbf{Inference}: Decisions made with respect to deployment and making predictions on unseen data inputs.
    \item \textbf{Tuner}: Once the above components have been defined, this component selects the top configuration and uses it for post-deployment model monitoring, retraining, and uncertainty quantification.
\end{itemize}

\subsection{Task formulation}
In time series forecasting, we are provided with a time series $X=(x_1,\dots,x_T)$, $x_i\in\mathbb{R}$ and a pre-specified prediction horizon $h\in\mathbb{N}$ as the objective. Then a quality metric $\mathcal{L}:\mathbb{R}^h\times\mathbb{R}^h\to\mathbb{R}$ is defined to evaluate the prediction quality of the forecasting model $M:\mathbb{R}^T\to \mathbb{R}^h$. The objective is to minimize the expectation of this metric $\mathbb{E}[\mathcal{L}(M(X), X^+)]$, where $X^+=(x_{T+1},\dots,x_{T+h})$ is a sequence representing the ground truth future values.

\subsection{Data Aspects}
\label{sec:data}
Data is a critical component machine learning applications and shapes most decisions made within the pipeline. Real-world time series datasets often contain missing values, noise, and do not follow regularities seen in many benchmark datasets. Most prior works consider the data to be \textit{static} and improvements are solely sought at the model/algorithmic level. We instead consider data modifications as part of the overall modelling effort when making comparisons.

Extending the time series definition, the dataset $\mathcal{D} = (X, C)$ consists of a set of $n$ series $X=\{X^1,\dots,X^n\}$, with $X^i = (x_1^i, x_2^i, \dots, x_T^i)$, where $T$ is the length of the time series. Accompanying $X$ is a set of covariates $C=\{C^1,\dots,C^n\}$, $C^i = (c_1^i, c_2^i, \dots, c_T^i)$. Here, the time series and covariates can be vectors of arbitrary dimension, i.e. $x_t^i \in \mathbb{R}^d$ and $c_t^i \in \mathbb{R}^k$, where $d$ and $k$ are the dimensions of the time series and covariates, respectively. For the case when $d=1$ the time series is considered \textit{univariate} and when $d>1$ it is considered \textit{multivariate}.

In this way of defining the dataset multiple assumptions can be violated 1) \textit{Series in $X$ are independent of each other} - in this case a static data $S$ may be present within the dataset $\mathcal{D} = (X, C, S)$ which can contain correlation information between the series (e.g. spatial location of the sensors which were used for data collection) 2) Series in $X$ are of the same length - in this case the length $T$ can vary between series  3) \textit{Prior knowledge about the future is lacking where in fact it could indeed be contained within the covariates or a variable.} - in this case covariates $C$ are split into two disjoint sets: $C_o$ which contains covariates that we have no future knowledge on and $C_k$ where we do have such knowledge. For example we may know an event occurring in the future in advance or we have full control over a variables.

Using the dataset $\mathcal{D}$, the \textit{data-split} module generates the splits $(D_{\textrm{train}}, D_{\textrm{val}}, D_{\textrm{test}})$ corresponding to the train, validation and test sets, respectively, and is fixed amongst methods. Beyond this point, how the data is shaped can dramatically effect how it can be consumed by downstream models. For example, an underlying method may be sensitive to the dynamic range and may require normalization, or on other hand cannot accommodate covariates as additional input, hence these must be excluded. The pre-processing function $p$ implements a variety of functionalities, such as filtering, outlier removal, imputation, label correction, feature selection, and so on. It may also be the case that at each pre-processing step multiple viable alternatives can be applied to the data to improve its \textit{quality} for a particular model and should be exposed to a tuner when evaluating the overall performance. In general $p$ is set to satisfy certain modeling conditions and downstream assumption. 

\subsection{Training Aspects}
Once the dataset is prepared, we must determine how much exploration is necessary for model training and how to ensure reliable training. In general, a forecasting model $M$ is a function which takes a series of observed values as well as the covariates as inputs, and predicts a series of forecast outputs:
\begin{align}
    M(S, X, C) \mapsto \hat{X}^+ \in \mathbb{R}^{d\times h}
\end{align}

Due to finite computational resources, it's common to limit the data used for training. This is achieved through a look-back window, which varies across models. A standard assumption is that the most recent values of a series contain the most relevant prior for future predictions. A fixed-length look-back window $l$ is used, starting at the index where future predictions are desired. More concretely, let $X'$ be the set of series from $X$ trimmed to the last $l$ observations, i.e. $X':= \{(x^i_{T-l+1},\dots, x^i_T) |  (x^i_1,\dots x^i_T)\in X\}$ and analogously define $C_k'$ and $C_o'$. Following this assumption, the history is approximated using the look back window:

\begin{align}
M(S,X,C_k,C_o) \approx M(S,X',C_k',C_o')
\end{align}

Here, the model $M$ can either operate at a \textit{global} level in which model parameters are estimated using all available time series in the dataset or on the other extreme at a \textit{local} level where a single time series in the dataset is used to estimate model parameters \citep{monash}. The choice of $M$ may also result in different training parameters which can include optimization parameters, initialization, objectives, whether to apply curriculum learning, exponential moving averaging (EMA) and so on. These sets of parameters must also be considered jointly with model exploration as will be discussed in Sec, \ref{sec:tuner}. Thus it is important to consider parameter ranges that control model complexity and training complexity.

The quality of the model $M$ is measured by evaluating model's predictions given ground truth values using a loss function
\begin{align}
    \mathcal{L}:\mathbb{R}^{d\times h}\times \mathbb{R}^{d\times h} \to \mathbb{R}
\end{align}

Here, often we assume \textit{there is only one metric that is useful} to serve for both as a training objective and for model selection, which may not be the case. The stopping criterion used to terminate training early can also effect final model selection. Having pre-specified dataset decisions, in the training portion the objective is then to find the best data-informed model that has the lowest loss on a held out dataset.

\subsection{Inference Aspects}

Successful models often rely on certain assumptions about the series they aim to predict, such as stationarity or no gaps in the data. Generally the forecast window predicted by the model is set by the downstream task, resulting in predictions $X^i_+ = (x^i_{T+1}, \dots, x^i_{T+h})$. Similar to how a preprocessing function $p$ was defined in Sec. \ref{sec:data} to transform the raw series, a post-processing function $p'$ operates on the forecast window to reverse certain aspects of the preprocessing decisions steps (for example, normalization) and incorporate domain expert knowledge about how the predicted window should be interpreted (for example, product sales cannot be negative and setting a lower bound).

Successful models often rely on certain assumptions about the data they aim to predict, such as stationarity and no gaps in the series. These assumptions simplify the modeling process and allow for more accurate predictions. However, real-world data rarely adheres perfectly to such assumptions. To compensate for this, similar to preprocessing techniques are often employed to transform the raw data into a format that satisfies the model's output requirements.

The forecast window predicted by the model is typically determined by the downstream task it serves. For example, a sales forecasting model might be designed to predict product demand for the next month, resulting in a forecast window $X^i_+ = (x^i_{T+1}, \dots, x^i_{T+h})$, where $T$ is the current time and $h$ is the prediction horizon.

While preprocessing prepares the data for modeling, a post-processing step ensures the predicted values are interpretable and aligned with domain knowledge. This post-processing function, denoted as $p'$, operates on the forecast window and performs several key tasks. It reverses certain aspects of the preprocessing steps, such as denormalization, to return the predicted values to their original units. Additionally, it incorporates domain-specific knowledge to ensure the predictions are realistic and consistent with real-world constraints. For instance, a post-processing function for product sales might enforce a non-negativity constraint, preventing the model from predicting negative sales figures.

By combining preprocessing and post-processing steps, we can leverage the strengths of machine learning models while ensuring the predictions are interpretable and applicable to the specific domain.

\subsection{Tuner Hyperparameter Selection}
\label{sec:tuner}

In the aforementioned sections we have considered a series of hyperparameter configurations spaces which must be explored to ensure reliable model training and comparison. There is a little incentive for authors to reproduce other paper's results in their training setup, and this in turn may put the reference models at a disadvantage. To ensure a fair comparison models should be implemented within the same framework with access to the same modifications across the pipeline as well as hyperparameter optimization (HPO) algorithms. Let $\Lambda = \Lambda_1 \times \Lambda_2 \times \cdots \Lambda_n$ define the space of all hyperparameters, and a vector in this hyperparameter space be denoted by $\lambda \in \Lambda$. Within this set of hyperparameters a series of ranges for each are considered, and is denoted by $\mathcal{A}$. The hyperparameters consists of both model specific $\lambda_{M}$ and dataset or training specific parameters $\lambda_{P}$, that is $\lambda = [\lambda_{M};\lambda_{P}]$. Then the objective of the tuner is to find 

\begin{align}
    \lambda_{top} = \argmin_{\lambda \in \mathcal{A}} \mathbb{E}_{(D_{train} \cup D_{val}, D_{test}) \sim \mathcal{D}} V(\mathcal{L}, \lambda, D_{train} \cup D_{val}, D_{test}), 
\end{align}

where $V$ measures the loss of the framework set with hyperparameters $\mathcal{A}_\lambda$. When running this tuner for each model, a fixed budget $B$ for finding the best configuration defined in terms of compute hours is used when running the pipeline on the same system and environment. In order to obtain the final model $M$ a two-fold optimization process is used: 1) a hyperparameter $\lambda$ is sampled from $\mathcal{A}$ 2) an optimization procedure is applied on model weights using $D_{train} \cup D_{val}$. The model optimization procedure terminates when it meets a stopping condition evaluated on a validation set. 2) step 1 is repeated until the HPO budget $B$ is exhausted. In the next section we explain how models protability and generalizability is assessed. The process is depicted with more detail in the flow chart depicted in Figure \ref{fig:pipeline_overview} and Section \ref{app:flow} of the Appendix.

\subsection{Model Selection \& Evaluation}

In order to compare models, simply evaluating a single trained model is not adequate due to the inherent variance in the performance for such tasks. For example, simply changing random seeds will result in an accuracy spread which may be deemed statistically significant as done in \citep{picard2021torch}. Instead, the top configuration found by the tuner is used to initialize the family of parametric models $q(\theta | \mathcal{D}, \lambda_{top})$, from which model are sampled and trained. The trained model is denoted as $p(y|\mathcal{D}, \theta))$. We sample $K$ such models $\theta_m \sim q(\theta | \mathcal{D}, \lambda_{top})$, and evaluate using the loss:

\label{eq:mean_loss}

\begin{align}
    \mathbb{E}[\mathcal{L}(p(y|\mathcal{D};\theta), y)] \approx \frac{1}{K} \sum_{m=1}^{K}\mathcal{L}(p(y| \mathcal{D}, \theta_m), y)
\end{align}

Similarly, the variance of the losses is computed and represented as $\sigma^2 = \text{Var}(\mathcal{L}(p(y|\mathcal{D};\theta), y))$. To compare two models a statistical test is formulated to determine if model $M_1$ has a lower expected loss than the other model $M_2$. Let $q_1$ be the parametric family of models for $M_1$, along with its expected loss $E_1$, and variance $\sigma_1^2$. Similarily, $q_2$ be the parametric model for $M_2$, $E_2$ the expected loss and $\sigma_2^2$ its variance. The null hypothesis posits $h_0$: \textit{there is no significant difference between the expected losses of the models}. Then the $t$ value is given by 

\begin{align}
    t  = \frac{E_1 - E_2}{\frac{1}{K}\sqrt{\sigma_1^2 + \sigma_2^2}}
\end{align}

By specifying a desired significance level $\alpha$, the null hypothesis is rejected if $t > t_{crit}$ where $t_{crit}$ is taken from the $t$-distribution table. In the case where there is a significance difference it can easily be determined which model is better by picking the one with lower expected loss. This method of comparing is robust to random initialization and training settings, which also better reflects the portability of the model across datasets. 

\section{Experiments}
\subsection{Dataset Details}

\begin{table}[h]
\label{tab:dataset_summary}
\centering
\caption{Summary of datasets.}
\begin{tabular}{lcccc}
    \toprule
    & Electricity & PEMS-BAY & M5 & Wikitraffic \\
    \midrule
    Num. Series ($n$) & 370 & 325 & 30490 & 145063 \\
    Sample Rate & 1h & 5 min & 1 day & 1 day \\
    Look-back Window ($l$) & 168 & 12 & 28 & 64 \\
    Horizon Length ($h$) & 24 & 12 & 28 & 64 \\
    Covariates ($C$) & \xmark &  \cmark & \xmark & \xmark \\
    Static ($S$) &  \xmark & \cmark & \cmark & \cmark \\
    \bottomrule
\end{tabular}
\end{table}

Time series model benchmarking can be challenging due to the diversity of data characteristics. Unlike in natural language processing (NLP) or computer vision (CV), where data tends to have more consistent characteristics, time series data can vary significantly in terms of its structure and complexity\citep{dataset_survey}. This makes it difficult to accurately compare the performance of different models and methodologies, as there may not be a one-size-fits-all approach that works well across all datasets. The data-centric approach taken considers the variations in how the data is transformed as an additional tunable parameter.

In our experiments, four widely used datasets in time series forecasting benchmarking are considered and summarized in Table. \ref{tab:dataset_summary}: the Electricity, PEMS-BAY, Wiki-Traffic Prediction, and the M5 dataset. For more details on each dataset refer to Appendix \ref{app:dataset}

For these datasets we consider various normalization preprocessing steps $p$: $\textrm{Z}(x) := \frac{x - \mu}{\sigma}$ where $\mu$ is the mean and $\sigma$ the standard deviation, log transformation $\textrm{log}(x) := \text{log}(x+1)$, and the composite of these $\textrm{Z}(\textrm{log}(x)) := \textrm{Z}(\textrm{log}(x+1))$.

Feature-engineering is also considered as a preprocessing step on the input which is detailed in Appendix \ref{app:dataset} and depends on the dataset properties. For preprocessing functions that require fitting on an underlying data, they are done so on the $\mathcal{D}_{\text{train}}$. The validation $\mathcal{D}_{\text{val}}$ and test  $\mathcal{D}_{\text{test}}$ data splits are standardized based on the training split of each dataset.

\subsection{Models Details}

For our experiments, the considered models contain well-established reference implementations and demonstrate state-of-the-art performance on the datasets listed in the previous section. These models are re-implemented and optimized within the proposed framework, achieving performance that either matches or surpasses the reference implementations. The summary of these models is presented in Table \ref{tab:models}.

\begin{table}
    \label{tab:models}
    \centering
    \caption{Table summarizing models used in experiments, and their operating mode. In the table, RNN $:=$ Recurrent Neural Networks, CNN $:=$ Convolutional Neural Networks, GNN $:=$ Graph Neural Networks.}
    \begin{tabularx}{\textwidth}{llllll}
    \toprule
        \textbf{Model} & \textbf{Summary} &  \textbf{Univariate} & \textbf{Multivariate} & \multicolumn{2}{c}{\textbf{Inputs}}\\
        \cmidrule(r){5-6}
        & & & & covariates & static \\
    \midrule
        ARIMA \citep{arima} & Statistical Model & \cmark & \xmark & Yes & No \\
        DeepAR \citep{deepar} & RNN  & \cmark & \xmark & Yes & Yes \\
        TFT \citep{tft} & Transformer+RNN &  \cmark & \cmark & Yes & Yes \\
        N-BEATS \citep{nbeats} & MLP  & \cmark & \xmark & No & No \\
        N-HITS \citep{nhits} & MLP  & \cmark & \xmark & No & No \\
        MTGNN \citep{mtgnn} & CNN + GNN & \xmark & \cmark & Yes & Yes \\
        DCRNN \citep{dcrnn} & GNN + RNN  & \xmark & \cmark & Yes & Yes \\
        XGBoost \citep{chen2015xgboost} & Boosting Decision Trees  & \cmark & \cmark & Yes & Yes \\
    \bottomrule
    \end{tabularx}
\end{table}
    
Each model has a corresponding set of hyperparameters $\lambda_m$ which pertain to the hidden dimension sizes, number of layers, etc, the complete list and the ranges considered for each dataset and model is specified in Appendix \ref{app:model_hpo}. Note that amongst these models is also a strong baseline $\textrm{XGBoost}$, often used to compete in competitions with additional feature engineering pre-processing steps.

\subsection{Evaluation Metrics}
The following metrics are considered when evaluating each model:
\begin{itemize}
    \item Mean Absolute Error (MAE): $\frac{1}{n}\sum^n|X^+ - \hat{X}^+|$ 
    \item Root Mean Squared Error (RMSE): $\sqrt{\frac{1}{n}\sum^n(X^+ - \hat{X}^+)^2}$ 
    \item Symmetric mean absolute percentage error (SMAPE): $\frac{1}{n}\sum^n\frac{|X^+ - \hat{X}^+|}{(|X^+| + |\hat{X}^+|)/2}$ * 100
    % \item Temporal Distortion Index (TDI): this metric is calculated by computing dynamic time warping (DTW) \citep{muller2007dynamic} metric on the forecast and ground truth, normalized by the sequence length.
\end{itemize}
Each metric is computed pointwise regardless of the forecast window $h$. Many publications use MAPE\citep{gman,st-wavenet,mtgnn,graph_wavenet} or q-risk\citep{tft,deepar} metrics, and since MAPE is unbounded (if the ground truth approaches 0) we chose to use SMAPE instead.

\subsection{Training Loop \& Optimizers}
Irrespective of models and datasets, there are techniques that can be leveraged when training time series forecasting models to improve performance. In particular the choice of optimizer, for example Adam \citep{kingma2014adam} and its hyperparameters such as initial learning rate, momentum, etc. Techniques that directly modify the training loop like exponential move averaging (EMA), where an offline model keeps track of the model-weights using a exponential moving average similar to \citep{grill2020bootstrap, izmailov2018averaging} or curriculum learning (CL) \citep{bengio2009curriculum} are also considered to be tunable.

\subsection{Tuner \& Compute Details}

The hyperparameters considered for the tuner to optimize are summarized in Table \ref{tab:hpo_space} and their corresponding ranges setting the subspace of valid hyperparameters $\mathcal{A}$.

All our experiments are conducted on DGX1-32G system with 8xV100 GPUs.  The computational budget $B$ is set to one DGX week i.e. 1344 GPU hours for each model-dataset combination. For evaluating models, the top configuration is trained $K=256$ times for Electricity and PEMS and $K=64$ for M5 and Wikitraffic datasets, and then evaluated on the test set using the metrics described in the section 4.3.

\begin{table}
    
    \centering
    \caption{Summary of hyperparameters considered specific hyper-parameters for various models and a more extensive list is provided in the Appendix \ref{app:model_hpo}. \label{tab:hpo_space}}
    \begin{tabularx}{\textwidth}{llrrrrr}
    \toprule
    Component & Parameters & \multicolumn{4}{c}{Values} \\
    \cmidrule(r){3-6}
    & & Electricity & PEMS-BAY & WikiTraffic & M5 \\
    \midrule
    \multirow{3}{1em}{Dataset} & \tiny Norm & \tiny $\textrm{Z}$ & \tiny $\textrm{Z}$ & \tiny [$\textrm{Z}$, $\textrm{log1p}$, $\textrm{Z(log1p)}$] & \tiny [$\textrm{log1p}$, $\textrm{Z(log1p)}$] \\
    & \tiny Covariates & \tiny F & \tiny T/F & \tiny F & \tiny F \\ 
    & \tiny Static & \tiny F & \tiny T/F & \tiny T/F & \tiny T/F \\ 
    % & \tiny look back window & \tiny F & \tiny T/F & \tiny F & \tiny F \\ \hline
    \hline
    \multirow{4}{1em}{Training} & \tiny CL & \tiny T/F & \tiny T/F & \tiny T/F & \tiny T/F \\
    & \tiny EMA & \tiny T/F & \tiny T/F & \tiny T/F & \tiny T/F \\
    & \tiny EMA decay & \tiny (0.9, 0.9999) & \tiny (0.9, 0.9999) & \tiny (0.9, 0.9999) & \tiny (0.9, 0.9999) \\
    & \tiny Optimizer & \tiny [Adam] & \tiny [Adam] & \tiny [Adam] & \tiny [Adam] \\
    & \tiny Criterion & \tiny [L1, GLL, MSE] & \tiny [L1, GLL, MSE] & \tiny [L1, GLL, MSE] & \tiny [L1, GLL, MSE, Tweedie] \\
    & \tiny Learning Rate & \tiny (1e-5, 1e-2) & \tiny (1e-5, 1e-2) & \tiny (1e-5, 1e-2) & \tiny (1e-5, 1e-2) \\
    & \tiny Dropout & \tiny (0, 0.5) & \tiny (0, 0.5) & \tiny (0, 0.99) & \tiny (0, 0.9) \\
    %& \multirow{2}{1em}{\tiny Criterion} & \tiny{[MAE, MSE,} & \tiny{[MAE, MSE,} & \tiny {[MAE, MSE,} & \tiny{[MAE, MSE,}  \\ 
    %& & \tiny{SMAPE, TDI]} & \tiny{SMAPE, TDI]} & \tiny {SMAPE, TDI]} & \tiny{SMAPE, TDI]} \\\hline
    \multirow{1}{1em}{Inference} & \tiny post-processing & \tiny - & \tiny - & \tiny clip to zero and round & \tiny clip to zero and round \\
    % & \tiny 40 & \tiny 13 & \tiny 14 & \tiny 15 & \tiny 16 \\
    \bottomrule
    \end{tabularx}
\end{table}

\section{Results}

% performs best across the datasets. 
Table \ref{tab:results} summarizes the empirical performance of the different models on the target datasets within the proposed framework and experimental setup. Notably, deep learning models exhibit strong competitive performance against XGBoost models, which leverage selective feature engineering and expert knowledge, on several individual datasets. This finding challenges the assumption that traditional, feature-engineered models consistently outperform deep learning approaches in time series forecasting tasks. Moreover, the results showcase the absence of a single model that reigns supreme across all datasets, highlighting the importance of considering data-specific characteristics when selecting a forecasting model.

Building upon the individual model performance, we explored ensemble learning, a widely recognized strategy to boost forecasting accuracy in time series tasks \citep{ensembling}. To this end, we trained up to 256 instances of each model with optimized hyper-parameters derived from different random seeds, resulting in a diverse ensemble of predictors. The final prediction was obtained through decision fusion via prediction averaging, leveraging the collective wisdom of the ensemble; Table \ref{tab:ensemble} presents the summary of results for the ensemble models. Through the experiments conducted on these models a series of observations can be made:

\textbf{Metrics}: Different models can be superior depending on the chosen metric, emphasizing the importance of considering a collective of metrics for model comparison. We excluded the TDI metric as the tuner was unable to effectively utilize it for model discrimination. This highlights the need of including multiple evaluation metrics that are both informative and readily usable by automated optimization tools.

\textbf{Effect of Batch Size}: Contrary to domain-specific recommendations favoring smaller or larger batch sizes, our findings suggest a tighter coupling between batch size and the specific dataset and model. For example, DeepAR prefers smaller batch sizes, while N-BEATS exhibits insensitivity to this hyperparameter, and TFT requires a specific range.

\textbf{CL \& EMA}: In contrast to previous claims, the investigated training techniques did not consistently improve performance across datasets. Notably, curriculum learning (CL) provided negligible improvement despite efforts to encourage the tuner to enable it. Exponential moving averaging (EMA) produced mixed results, potentially due to the noise inherent in the forecasting task and the detrimental effects of EMA smoothing.

\textbf{Context Length (l)}: The optimal context length exhibits a strong dependency on the specific model and dataset combination. This implies that a single, universal context length recommendation is unlikely to be effective across all models and datasets. N-BEATS and NHITS, for example, demonstrate poor performance when provided with limited context (less than 256 data points). However, their performance significantly improves with access to a richer context (greater than 256 data points).

\textbf{Weak Models as Strong Baselines in Ensembles}: DeepAR, despite its subpar performance when trained individually (e.g., on the WikiTraffic dataset), exhibits a remarkable transformation when incorporated into ensembles. As shown in Table \ref{tab:ensemble}, ensembles utilizing DeepAR achieve significantly superior performance compared to other models. This intriguing finding underscores the potential of seemingly weak models to act as powerful baselines within ensemble configurations. By leveraging the diverse strengths of individual models, ensembles can collectively achieve superior performance, even when some components under-perform in isolation.

\begin{table}
    \centering
    \caption{This table presents the mean error performance of various forecasting models, along with their corresponding standard deviations. Lower values indicate better performance.\label{tab:results}}
    \resizebox{\textwidth}{!}{%
    \begin{tabular}{ l *{12}{c} }
    \toprule
    & \multicolumn{3}{c}{Electricity} & \multicolumn{3}{c}{PEMS-BAY} & \multicolumn{3}{c}{WikiTraffic} & \multicolumn{3}{c}{M5} \\
    \cmidrule(r){2-4}\cmidrule(r){5-7}\cmidrule(r){8-10}\cmidrule(r){11-13}
    Model & MAE$\downarrow$ & RMSE$\downarrow$ & SMAPE$\downarrow$  & MAE$\downarrow$ & RMSE$\downarrow$ & SMAPE $\downarrow$ & MAE$\downarrow$ & RMSE$\downarrow$ & SMAPE$\downarrow$ & MAE $\downarrow$ & RMSE$\downarrow$ & SMAPE$\downarrow$ \\
    \midrule
    TFT & 41.72 $\tiny\pm$ 0.714 & 352.85 $\tiny\pm$ 12.74 & \textbf{8.65} $\tiny\pm$ 0.0572  & 1.64 $\tiny\pm$ 0.024 & 3.83 $\tiny\pm$ 0.072 & 3.34 $\tiny\pm$ 0.0573  & \textbf{308.4} $\tiny\pm$ 6.683 & 12343.06 $\tiny\pm$ 817.28 & \textbf{38.83} $\tiny\pm$ 0.48 & 1.07 $\tiny\pm$ 0.01 & 2.60 $\tiny\pm$ 0.09 & 146.88 $\tiny\pm$ 0.55 \\
    N-BEATS & 43.15 $\tiny\pm$ 0.227 & 341.65 $\tiny\pm$ 3.29 & 9.45 $\tiny\pm$ 0.0527  & 1.97 $\tiny\pm$ 0.001 & 4.67 $\tiny\pm$ 0.007 & 3.99 $\tiny\pm$ 0.0027  & 322.66 $\tiny\pm$ 14.38 & \textbf{12191.58} $\tiny\pm$ 1504.04 & 40.38 $\tiny\pm$ 0.41 & 1.02 $\tiny\pm$ 6e-4 & 2.46 $\tiny\pm$ 4e-3 & 159.90 $\tiny\pm$ 0.25 \\
    N-HITS & 42.30 $\tiny\pm$ 0.093 & 340.56 $\tiny\pm$ 1.51 & 9.25 $\tiny\pm$ 0.0146  & 1.95 $\tiny\pm$ 0.001 & 4.62 $\tiny\pm$ 0.005 & 3.92 $\tiny\pm$ 0.0023  & 332.30 $\tiny\pm$ 8.44 & 13554.27 $\tiny\pm$ 855.14 & 39.86 $\tiny\pm$ 0.18 & 1.01 $\tiny\pm$ 5e-4 & 2.46 $\tiny\pm$ 3e-3 & 159.91 $\tiny\pm$ 0.10 \\
    MTGNN & 43.00 $\tiny\pm$ 0.537 & 342.64 $\tiny\pm$ 6.56 & 9.44 $\tiny\pm$ 0.1032  & 1.55 $\tiny\pm$ 0.005 & 3.61 $\tiny\pm$ 0.014 & 3.11 $\tiny\pm$ 0.0107  & $-^*$ & $-^*$ & $-^*$ & $-^*$ & $-^*$ & $-^*$ \\
    DCRNN & $-^{**}$ & $-^{**}$ & $-^{**}$ & \textbf{1.54} $\tiny\pm$ 0.007 & \textbf{3.57} $\tiny\pm$ 0.0187 & \textbf{3.08} $\tiny\pm$ 0.0138  & $-^{**}$ & $-^{**}$ & $-^{**}$ & $-^{**}$ & $-^{**}$ & $-^{**}$ \\
    DeepAR & 46.98 $\tiny\pm$ 1.53  & 367.54 $\tiny\pm$ 28.86 & 10.27 $\tiny\pm$ 0.1546  & 1.73 $\tiny\pm$ 0.011 & 3.88 $\tiny\pm$ 0.0279 & 3.51 $\tiny\pm$ 0.0224  & 9108.69 $\tiny\pm$ 47456.31 & 10633457.81 $\tiny\pm$ 55977165.11 & 39.47 $\tiny\pm$ 0.38 & 1.01 $\tiny\pm$ 3e-3 & 2.47 $\tiny\pm$ 0.02 & 158.88 $\tiny\pm$ 0.65 \\
    XGBoost & \textbf{40.67} $\tiny\pm$ 0.03 & \textbf{327.68} $\tiny\pm$ 0.59 & 8.78 $\tiny\pm$ 2e-3 & 1.66 $\tiny\pm$ 2e-4 & 3.87 $\tiny\pm$ 5e-4 & 3.35 $\tiny\pm$ 4e-4 & 317.58 $\tiny\pm$ 0.46 & 12801.7 $\tiny\pm$ 85.2 & 38.93 $\tiny\pm$ 0.01 & \textbf{1.002} $\tiny\pm$ 4e-4 & \textbf{2.45} $\tiny\pm$ 2e-3 & \textbf{127.91} $\tiny\pm$ 0.086 \\
    \bottomrule
    \multicolumn{12}{p{0.8\textwidth}}{Notes: \*Experiments not performed due to memory requirements. \newline \*\*DCRNN requires static graph which is available only for PEMS-BAY}
    \end{tabular}%
    }
\end{table}

\begin{table}
    \label{tab:ensemble}
    \centering
    \caption{Error of different forecasting models using an ensemble of models the mean performance is reported for each. Ensemble of 256 model was used for Electricity and PEMS-BAY and ensemble of 64 models for Wikitraffic and M5.}
    \resizebox{\textwidth}{!}{%
    \begin{tabular}{ l *{12}{c} }
    \toprule
    & \multicolumn{3}{c}{Electricity} & \multicolumn{3}{c}{PEMS-BAY} & \multicolumn{3}{c}{WikiTraffic} & \multicolumn{3}{c}{M5} \\
    \cmidrule(r){2-4}\cmidrule(r){5-7}\cmidrule(r){8-10}\cmidrule(r){11-13}
    Model & MAE$\downarrow$ & RMSE$\downarrow$ & SMAPE$\downarrow$ & MAE$\downarrow$ & RMSE$\downarrow$ & SMAPE$\downarrow$ & MAE$\downarrow$ & RMSE$\downarrow$ & SMAPE$\downarrow$ & MAE$\downarrow$ & RMSE$\downarrow$ & SMAPE$\downarrow$ \\
    \midrule
    TFT & \textbf{40.40} & 340.04  & \textbf{8.43}  & 1.55 & 3.57  & 3.13  & 297.95 & 11896.13 & \textbf{37.86} & 1.07 & 2.58 & 146.83 \\
    N-BEATS & 42.57  & 338.70 & 9.27  & 2.03  & 4.46 & 4.11  & 303.87 & \textbf{10883.19} & 39.55 & 1.01 & 2.45 & 159.61 \\
    N-HITS & 41.71 & 338.17  & 9.06  & 1.94  & 4.60  & 3.91  & 320.81 & 12908.48 & 39.45 & 1.01 & 2.46 & 159.75 \\
    MTGNN & 41.71  & 331.11  & 9.20  & 1.52  & 3.54 & 3.05  & $-^*$ & $-^*$ & $-^*$ & $-^*$ & $-^*$ & $-^*$ \\
    DCRNN & $-^{**}$ & $-^{**}$ & $-^{**}$ & \textbf{1.47}  & \textbf{3.40}  & \textbf{2.95}  & $-^{**}$ & $-^{**}$ & $-^{**}$ & $-^{**}$ & $-^{**}$ & $-^{**}$ \\
    DeepAR & 42.34  & 329.03  & 9.58  & 1.67  & 3.68  & 3.36  & \textbf{292.07} & 11412.32 & 37.94 & \textbf{1.00} & \textbf{2.43} & 158.42 \\
    XGBoost & 40.44 & \textbf{325.58} & 8.73 & 1.65 & 3.85 & 3.33 & 317.31 & 12793.09 & 38.91 & 1.002 & 2.44 & \textbf{127.85} \\
    \bottomrule
    \multicolumn{12}{p{0.8\textwidth}}{Notes: *Experiments not performed due to memory requirements. \newline **DCRNN requires static graph which is available only for PEMS-BAY}
    \end{tabular}%
    }
\end{table}

\section{Conclusion}
The lack of standardization within the time series forecasting domain presents a significant impediment to progress. This issue manifests in inconsistent reporting practices, even when using the same dataset, as evidenced by discrepancies in metrics reported across various studies \citep{tft,mtgnn,yu2016temporal}. Additionally, the frequent reporting of results obtained with suboptimal hyper-parameters further obscures the true performance potential of different models, making it difficult to discerningly judge their relative strengths and weaknesses \citep{tft}. This work addresses these challenges by proposing a standardized workflow and framework that facilitates seamless integration of diverse datasets and training/inference logic. Furthermore, this framework transparently exposes decision-making processes throughout the workflow, enabling optimization tools to guide researchers and developers in their exploration. Our empirical evaluations reveal that deep learning methods demonstrate competitive performance against XGBoost models, challenging the traditional perception that feature-engineered approaches consistently dominate this domain. By promoting standardization and transparency, this work paves the way for more robust and reproducible research in the field of time series forecasting.

\cleardoublepage

\bibliographystyle{ACM-Reference-Format}
\bibliography{refs} % Entries are in the refs.bib file

%\begin{comment}
\cleardoublepage
\appendix

\section{Additional Experiments}

\subsection{Impact of Lookback Window Length}\label{sec:context_len}
In this section an additional study is conducted on the effect of the lookback window length $l$ on the final model performance. In particular, MTGNN and DCRNN are designed to work with large chunks of data comprising of many series, i.e. they operate in a local multivariate setting ($n=1,d=325$), which imposes a limit on either the batch size or lookback window. For the sake of comparison, we also train TFT, DeepAR, N-BEATS and N-HITS on PEMS-BAY with the same lookback window as MTGNN and DCRNN as in the results presented in Table \ref{tab:results}. The ultimate goal of time series forecasting task should be predicting the future values based on all the available history, regardless of the context length. To this end, in addition to the lookback window length of 12 we sweep it across  24, 48 and 288, then apply the same procedure we defined in section \ref{sec:tuner} for each model. The results for the best configuration is provided in Table. \ref{tab:context_len_hpo}. 

\begin{table}[h]
    \centering
    \caption{Summary of result for varying lookback window lengths. The top configuration is selected based on the mean average error (MAE) as the criterion.}
    \label{tab:context_len_hpo}
    
    \resizebox{\textwidth}{!}{%
    \begin{tabular}{ l *{12}{c} }
    \toprule
    & \multicolumn{3}{c}{Context 24} & \multicolumn{3}{c}{Context 48} & \multicolumn{3}{c}{Context 288}  \\
    \cmidrule(r){2-4}\cmidrule(r){5-7}\cmidrule(r){8-10}
    Model & MAE$\downarrow$ & RMSE$\downarrow$ & SMAPE$\downarrow$  & MAE$\downarrow$ & RMSE$\downarrow$ & SMAPE $\downarrow$ & MAE$\downarrow$ & RMSE$\downarrow$ & SMAPE$\downarrow$\\
    \midrule
    TFT & 1.566 & 3.68 & 3.165  & 1.583 & 3.693 & 3.202  & 1.64 & 3.795 & 3.355 \\
    N-BEATS & 1.937 & 4.632 & 3.912 & 1.909 & 4.571 & 3.848 & 1.74 & 4.145 & 3.545 \\
    N-HITS* & - & - & - & - & - & - & 1.824 & 3.994 & 3.721 \\
    MTGNN** & 1.534 & 3.566 & 3.074 &  1.549 & 3.617 & 3.113 & - & - & - \\
    DCRNN** & 1.528 & 3.564 & 3.063 &  1.512 & 3.564 & 3.028 & - & - & - \\
    DeepAR & 1.681 & 3.789 & 3.391 & 1.666 & 3.773 & 3.363 & 1.895 & 4.186 & 3.869 \\
    \multicolumn{9}{p{0.8\textwidth}}{Notes: *Due similarity of N-BEATS and N-HITS architectures and their performances on context length of 12 we skipped contexts 24 and 48. \newline **For DCRNN and MTGNN experiments for context length of 288 were not performed due the memory requirements to run these}
    \end{tabular}%
    }
\end{table}

It is worth mentioning that the longer the lookback window generally results in longer model training. This means that with the same computational budget $B$, the HPO algorithm might not find an optimal configuration for longer lookback windows hence there is an inherent trade-off. The results presented in \ref{tab:context_len_hpo} are the results for the best configuration found by HPO. These are biased towards lower errors due to the stochastic nature in training these models. A stability test on the best configurations is conducted, summarized in Table. \ref{tab:context_len_stability} using a sample $>$ 100 models. Additionally, the best configuration for the XGBoost model which is manually crafted by a domain expert Kaggle Grandmaster, which uses a much larger lookback window comparatively to other models.

\begin{table}[h]
        \centering
        \caption{Error mean and variance for the best configurations found by HPO using a sample $>$ 100 models.}
        \label{tab:context_len_stability}
        % \resizebox{\t}{!}{%
        \begin{tabular}{ l *{12}{c} }
        \toprule
        Model & Lookback window length & MAE $\downarrow$ & RMSE $\downarrow$ & SMAPE $\downarrow$ \\
        \midrule
        % TFT & 24 & 1.639 \tiny$\pm$ 0.0103 & 3.878 \tiny$\pm$ 0.1341 & 3.33 \tiny$\pm$ 0.1056 \\
        TFT & 48 & 1.627 \tiny$\pm$ 0.0453 & 3.838 \tiny$\pm$ 0.1272 & 3.303 \tiny$\pm$ 0.1044 \\
        N-BEATS & 288 & 1.739 \tiny$\pm$ 0.0018 & 4.139 \tiny$\pm$ 0.0097 & 3.543 \tiny$\pm$ 0.0037 \\
        N-HITS & 288 & 1.723 \tiny$\pm$ 0.0016 & 4.109 \tiny$\pm$ 0.0079 & 3.511 \tiny$\pm$ 0.0036 \\
        MTGNN & 24 & 1.544 \tiny$\pm$ 0.0058 & 3.584 \tiny$\pm$ 0.0147 & 3.097 \tiny$\pm$ 0.0118 \\
        DCRNN & 48 & \textbf{1.53} \tiny$\pm$ 0.0045 & \textbf{3.562} \tiny$\pm$ 0.015 & \textbf{3.063} \tiny$\pm$ 0.0093 \\
        DeepAR & 48 & 1.685 \tiny$\pm$ 0.0186 & 3.825 \tiny$\pm$ 0.0419  & 3.398 \tiny$\pm$ 0.0369  \\
        XGBoost & 4043 & 1.577 \tiny$\pm$ 0.0012 & 3.685 \tiny$\pm$ 0.0043 & 3.216 \tiny$\pm$ 0.0026 \\
        \end{tabular}%
        % }
    \end{table}
% }
\subsection{Effect of Exponential Moving Average and Curriculum Learning}
In our sweeps we consider EMA and CL as training techniques that can be turned on or off when searching for the best model configuration. Even though TPE algorithm holds an internal state containing information about how beneficial particular hyperparameter are, it is hard to interpret this state in a meaningful way. Instead we summarize which of the top configurations selected this technique for the final model in Table. \ref{tab:ema}. In general EMA was found to be quite useful for improving downstream performance.

\begin{table}
    \centering
    \caption{EMA usage amongst top configurations.}
    \label{tab:ema}
    \resizebox{\textwidth}{!}{%
    \begin{tabular}{ l *{12}{c} }
    \toprule
    Model & Electricity & PEMS-BAY & PEMS-BAY 24 & PEMS-BAY 48 & PEMS-BAY 288 & M5 & Wiki Traffic\\
    \midrule
    TFT     & \cmark & \cmark & \cmark & \cmark & \cmark & \cmark & \cmark \\
    N-BEATS & \cmark & \cmark & \cmark & \xmark & \cmark & \xmark & \xmark\\
    N-HITS  & \cmark & \cmark & - & - & \cmark & \cmark & \xmark \\
    DeepAR  & \cmark & \cmark & \cmark & \cmark & \cmark & \cmark & \cmark \\
    DCRNN   & \cmark & \cmark & \cmark & \cmark & - & - & - \\
    MTGNN   & \cmark & \xmark & \xmark & \xmark & - & - & - \\
    \multicolumn{8}{p{1.0\textwidth}}{Notes: *Due similarity of N-BEATS and N-HITS architectures and their performances on context length of 12 we skipped contexts 24 and 48. \newline **For DCRNN and MTGNN experiments for context length of 288 were not performed due the memory requirements to run these}
    \end{tabular}%
    }
\end{table}

 For the case of curriculum learning we considered this method for MTGNN which was used explicitly by the authors in their proposed approach. As a result, sweeps were conducted on this hyperparameter for the MTGNN model. From the results summarized in \ref{tab:mtgnn_cl}, we found that on both electricity and PEMS-BAY HPO algorithm tends to choose configurations with lower curriculum learning update steps, which reflects that there is a tendency to for the HPO procedure to turn off this feature.

\begin{table}
    \centering
    \caption{Best configurations found by HPO algorithm on MTGNN. Baseline is the mean error value for the best configuration without CL, reported in the main section of the paper. }
    \label{tab:mtgnn_cl}
    % \resizebox{\textwidth}{!}{%
    \begin{tabular}{ l *{12}{c} }
    \toprule
    & \multicolumn{3}{c}{Electricity} & \multicolumn{3}{c}{PEMS-BAY} \\
    \cmidrule(r){2-4}\cmidrule(r){5-7}
    Configuration & MAE$\downarrow$ & RMSE$\downarrow$ & SMAPE$\downarrow$ & MAE$\downarrow$ & RMSE$\downarrow$ & SMAPE$\downarrow$ \\
    \midrule
    MTGNN w/o CL   & 41.57 & 324.43 & 9.33 & 1.55 & 3.59 & 3.10 \\
    MTGNN + CL & 41.85 & 329.34 & 9.38 & 1.55 & 3.57 & 3.10 \\
    %MTGNN - baseline  & 43.00 \tiny$\pm$ 0.537 & 342.64 \tiny$\pm$ 6.56 & 9.44 \tiny$\pm$ 0.1032  & 1.55 \tiny$\pm$ 0.005 & 3.61 \tiny$\pm$ 0.014 & 3.11 \tiny$\pm$ 0.0107 \\
    \end{tabular}%
    % }
\end{table}

\subsection{Performance analysis}
In Table \ref{tab:perf_analysis} the average time required to train each of the top performing models is provided. The time is calculated with respect to the average number of epochs required to converge to the best checkpoint. We observed that TFT consistently requires less time to train than XGBoost, while achieving comparable errors.

\begin{table}[h]
    \centering
    \caption{Average time required to train top performing configurations.}
    \label{tab:perf_analysis}
    % \resizebox{\textwidth}{!}{%
    \begin{tabular}{l *{4}{c}}
    \toprule
     & \multicolumn{2}{c}{Electricity} & \multicolumn{2}{c}{PEMS-BAY} \\
     % & \multicolumn{2}{c}{WikiTraffic} & \multicolumn{2}{c}{M5} \\
    \cmidrule(r){2-3}\cmidrule(r){4-5}
    %\cmidrule(r){6-7}\cmidrule(r){8-9}
     Model & convergence epoch & training time[s] & convergence epoch & training time[s] \\
     %& convergence epoch & training time[s] & convergence epoch & training time[s] \\
    \midrule
    TFT & 2.98 & 700.37 & 9.43 & 1912.27 \\
    %& 3.27 & 1465.61 & 4.86 & 1949.0 \\
    NBEATS & 15.78 & 570.73 & 19.27 & 1673.75 \\
    %& 10.20 & 3023.06 & 21.28 & 2051.49 \\
    NHITS & 22.85 & 824.58 & 24.96 & 2227.26 \\
    %& 7.31 & 1283.22 & 17.47 & 1752.85 \\
    MTGNN & 18.15 & 2386.161276 & 66.77 & 9841.90 \\
    %& -$^*$ & -$^*$ & -$^*$ & -$^*$ \\
    DCRNN & -$^{*}$ & -$^{*}$ & 33.71 & 11078.38 \\
    %& -$^{**}$ & -$^{**}$ & -$^{**}$ & -$^{**}$ \\
    XGBoost & - & 1989.19 & - & 2188.38 \\
    %& - & -$^{***}$ & - & -$^{***}$ \\
    DeepAR & 1.07 & 1448.98 & 8.84 & 1303.71 \\
    %& 4.13 & 4267.42 & 10.94 & 10522.19 \\
    \bottomrule
    \multicolumn{4}{p{0.8\textwidth}}{Notes: *DCRNN requires static graph which is available only for PEMS-BAY. 
    %*Experiments not performed due to memory requirements. \newline 
    %\newline ***Experiments are computationally incomparable with the rest.
    } 
    \end{tabular}%
    % }
\end{table}

\section{Framework Flowchart}
\label{app:flow}
The high sensitivity of models to hyperparameter changes and the high variability of accuracy require us to be rigorous in our experiments. To ensure a fair comparison, certain parts of the experimental flow must be fixed (train/val/test splits, metrics, hyperparameter optimization algorithm). Others, however, must be sampled by the hyperparameter optimization algorithm using the hyperparameter space $\mathbb{S}$: model architecture parameters, training loop, pre- and post-processing algorithms.

\begin{figure}[h]
  \centering
  \includegraphics[width=0.8\linewidth]{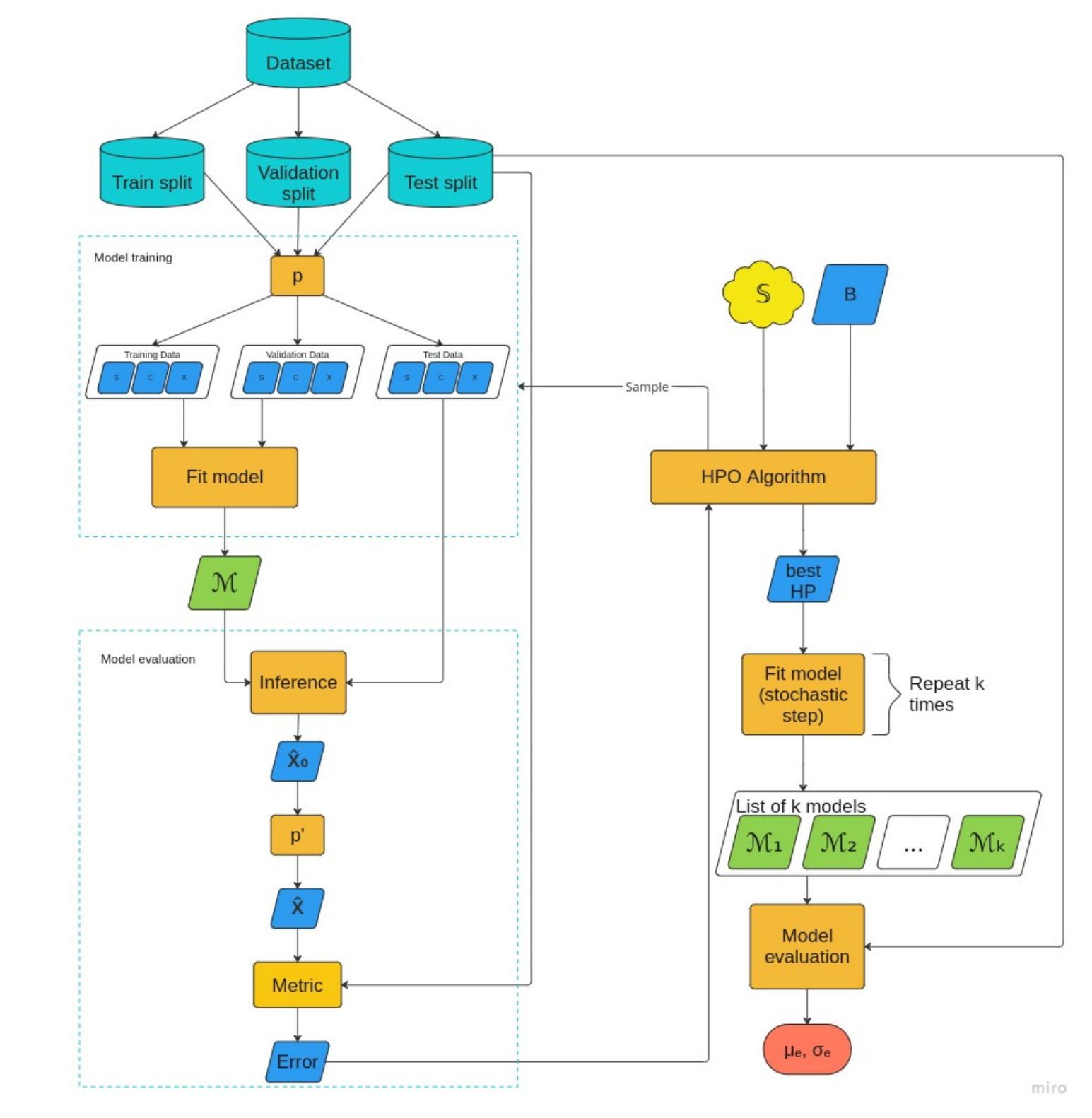}
  % \fbox{\rule[-.5cm]{0cm}{4cm} \rule[-.5cm]{4cm}{0cm}}
  \caption{A flowchart of the framework, where orange shows components that can be changed by a hyperparameter optimizatoin algorithm. }
  \label{fig:pipeline_overview}
\end{figure}

\section{Datasets}
\label{app:dataset}
Below we provide a more elaborate description for each dataset and how they are processed/standardized for the experiments.
\begin{enumerate}
%\subsection{Electricity}
\item{Electricity}

Electricity load diagrams\cite{electricity} (not to be confused with Electricity\cite{electricity_pricing}) is a commonly \cite{tft,deepar,yu2016temporal} used dataset. It consists of 369 distinct univariate time series without additional covariates.
For our experiments the data is processed as in \cite{tft} and \cite{deepar}, taking 500k randomly selected samples between 2014-01-01 to 2014-09-01 as a training set, from which the last week are used as validation split. The test set spans one week immediately after validation set, from 2014-09-01 00:00:00 to 2014-09-07 23:00:00. We compute a z-score on the training set (excluding validation subsplit) on each series separately and apply to each series separately. 

%\subsection{PEMS-BAY}
\item{PEMS-BAY}

The Caltrans Performance Measurement System (PeMS)\cite{pems} is a traffic measurement system spanning freeway system across all major metropolitan areas of the State of California. The system consists of more than 15000 sensors. In the literature one can find models evaluated on different subsets of this data. The most commonly used is PEMS-BAY as described \cite{dcrnn}.

%In DCRNN authors state "they selected 325 sensors in the Bay Area" and their selection criterion wasn't included in the paper. As for the road network graph, the authors state nothing more than that it was obtained using road distance. This specific version of the dataset was used in multiple subsequent publications making it pervasive in the literature\cite{gman,st-wavenet,stemgnn,mtgnn} . \\
For our experiments which require graph definition, we use the one precomputed by DCRNN authors, to closely match other papers \cite{gman,st-wavenet,stemgnn,mtgnn}. Each data point is a 5 minute aggregate of statistics of traffic collected by a single sensor. The objective is to predict the next hour (12 time steps) of traffic using previous hour as a prior. The dataset consists of the same 325 selected time series, where the training split ranges from 2017-01-01 to 2017-05-08, the validation split from 2017-05-08 to 2017-05-25 17:50:00, and the test split from 2017-05-25 17:50:00 to 2017-07-01. The first 12 steps of the test split are used as an encoder input to a model, thus labels range from 2017-05-25 18:50:00 to 2017-07-01.

%\subsection{M5}
\item{M5}
% The hierarchical structure allowed participants to use top-down, bottom-up or middle-out approaches in order to predict the lowest level of aggregation. Sparsity of the first 3 levels of aggregation, that is unit sales for each product in different locations, posed a significant challenge for participants, which could be observed in the population of 50 top performing teams, where they achieved at average 40\% more accurate predictions than benchmark on 1st (highest) level of aggregation, but only 3\% on the levels 10, 11 and 12. 

Kaggle M5\cite{m5} competition's objective was to accurately predict ~30.5k values of unit sales across different Walmart shops for the next 28 days. The dataset used for the competition has a hierarchical structure, where each level consists of entries aggregated by specific criterion, for example using same location, product or product category, etc. In total there were 42,840 series across 12 levels of aggregation. Evaluation was performed on all series, including aggregates, with the RMSE metric. In our experiments exactly the same dataset as in the M5 competition. We follow the standard train/valid/test splits: data from 2011-01-29 to 2016-04-24 for training set, from 2016-04-25 to 2016-05-22 for validation set and from 2016-05-23 to 2016-06-19 for test set. During our experiments an encoder length of 28 was used resulting in in example total length of 56. 
% Each target observation is processed with the $ln(1 + x)$ transform, so that resulting distribution resembles the Tweedie distribution \cite{tweedie1984index} that was used as a loss function in majority of top performing solutions during the contest. However, we noticed that DeepAR models tend to perform worse for unnormalized data, that's why for this model we also normalized the data after applying aforementioned transformation. For unnormalized dataset we define our loss function in terms of modified deviance from Tweedie distribution: $-\frac{y*\tilde{y}^{1-p}}{1-p} + \frac{\tilde{y}^{2-p}}{2-p}$ following LGBM implementation with parameter $p \in (1, 2)$ as a hyper parameter. For normalized one we used standard L1 loss, because tweedie loss is not defined for negative values. 
Due to the nature of our problem formulation, where we want to use one model to predict all the time series, we use bottom-up approach, trying to predict only series at level 12(lowest level of aggregation) and then aggregate predictions to produce higher levels without any correction.

%\subsection{Wiki traffic}
\item{Wiki traffic}\label{sec:wikitraffic}

Kaggle Web Traffic Time Series Forecasting \cite{wikitraffic} competition's objective was to predict next two months of daily 'hits' (visits) for 145k pages collected across 10 Wikimedia sites for different access and agent types. Authors choose to use SMAPE as a target metric to evaluate models' performance. The dataset consists of time series with different levels of popularity, which introduces an added complexity as models have to be robust to scale change, but even more significant is the fact that data for this challenge was collected as it came in and not all the pages from the original training set existed by the end of the competition, those pages were excluded from test evaluation. Additionally, original data contained 2 days gap between the validation and test set, due to the final train and validation sets being published 2 days before test data started being collected. That's why we used modified train/validation/test ranges, to account for the gap: for train set we used data points dated 2015-07-01 to 2017-07-08, for validation - 2017-05-06 to 2017-09-10 and for test - 2017-07-09 to 2017-11-13. Which, after the exclusion of 64 encoder inputs produces test set from 2017-09-11 to 2017-11-13 instead of from 2017-09-13 to 2017-11-13, as in the original competition data. We use longer prediction horizon of 64 instead of 62 to make dataset continuous, but we assign zero weight to the first 2 values in the test set to produce metrics comparable to those used during the competition. We remove zero values from the beginning of each series. Series that contain less than 64 observations in the training set after removing the zero prefix, are excluded from the dataset entirely.

\end{enumerate}

\section{Tuner}

In this section we provide more details on the specific ranges considered for each hyperparamter (i.e. specifiying $\mathcal{A}$ and the tuner setup when conducting the experiments \& evaluation. 

\subsection{Model Hyper-parameters}\label{app:model_hpo}
A summary of the hyperparameters used for various models is provded in Table \ref{tab:model_hpo_ranges}, where paranthesis "()" denote ranges, comma seperated values are explicit options, and brackets enclose model specific configurations.

\begin{table}[h]
    \label{tab:model_hpo_ranges}
    \centering
    \caption{Model specific hyper parameters and their ranges considered in this work.}
    \resizebox{0.74\textwidth}{!}{%
    \begin{tabular}{ l *{12}{c} }
    \toprule
    Model & Parameter name & Parameter range  \\
    \midrule
    TFT & \#head & 1, 2, 4 \\
    & hidden size & 96, 128, 192, 256 \\
    & dropout & (0, 0.5) \\
    \midrule
    N-BEATS & stack 0 type & trend, generic \\
    & stack 1 type & seasonality, generic \\
    & stack 0 \# blocks & 2, 4, 8 \\
    & stack 1 \# blocks & 2, 4, 8 \\
    & stack 0 theta dim & 2, 4, 8, 16 \\
    & stack 1 theta dim & 0, 2, 4, 8, 16 \\
    & stack 0 share weights & True, False \\
    & stack 1 share weights & True, False \\
    & stack 0 hidden size & 256, 512, 1024, 2048, 4096* \\
    & stack 1 hidden size & 256, 512, 1024, 2048, 4096* \\
    \midrule
    N-HITS & mlp layers & 2,3,4  \\
    & hidden size & 256,512,1024,2048 \\
    & activation & ReLU, Softplus, Tanh, SELU, LeakyReLU, PReLU, Sigmoid \\
    & pooling mode & MaxPool1d, AvgPool1d \\
    & \#blocks & [1,1,1],[1,1,2],[1,2,1],[1,2,2],[2,1,1],[2,1,2],[2,2,1],[2,2,2] \\
    & pool kernel size & [6,3,1],[6,2,1],[4,2,1],[3,3,1],[2,2,1] \\
    & freq downsample & [6,3,1],[6,2,1],[4,2,1],[3,3,1],[2,2,1] \\
    \midrule
    DeepAR & \# layers  & 2,3,4,5  \\
    & hidden size & 64,128, 256, 512, 1024, 2048** \\
    & dropout & (0, 0.6) \\
    \midrule
    MTGNN & gcn depth & 2,3,4 \\
    & dropout & (0,0.5) \\
    & subgraph size & 10,15,20 \\
    & node dim & 32,40,64,128 \\
    & conv channels & 16,32,64 \\
    & residual channels & 16,32,64 \\
    & skip channels & 32,64 \\
    & end channels & 32,64 \\
    & \#layers & 2,3,4 \\
    & propalpha & (0.01, 0.2) \\
    & tanalpha & (2.0, 4.0) \\
    & in dim & 16, 24, 32, 64 \\
    & embedding & True, False \\
    \midrule
    DCRNN & max diffusion step & 1, 2 \\
    & \#RNN layers & 2, 3 \\
    & RNN units & 32, 64, 128 \\
    & activation & tanh, ReLU \\
    \bottomrule
    \end{tabular}%
    }
\end{table}

\end{document}